\documentclass[preprint,5p,times,twocolumn,numbers]{elsarticle}

\usepackage{amssymb}
\usepackage{amsmath}
\usepackage{multirow}
\usepackage{hyperref}
\usepackage{url}
\usepackage[numbers]{natbib}

\usepackage{lineno} 
\usepackage{comment}

\journal{International Journal of Hydrogen Energy}

\begin{document}

\begin{frontmatter}
 
\title{How Mobile Gas Sensor Trajectories Govern Hydrogen Leak Detection:\\ A Safety Gap in Manual Leak Inspection of Hydrogen System Components}
 
\author[inst1]{Christian Masuhr}
\author[inst1]{Arne Wendt}
\author[inst1]{Thorsten Schüppstuhl}
 
\affiliation[inst1]{organization={Institute of Aircraft Production Technology (IFPT), Hamburg University of Technology (TUHH)},
            addressline={Denickestraße 17},
            city={Hamburg},
            postcode={21073},
            country={Germany}}
 
\begin{abstract}
The integrity of hydrogen infrastructure depends on reliable leak detection,
which in electrolyzer manufacturing is performed almost exclusively by manual
tracer gas sniffing. Although the procedure is mandated by international
standards, these constrain only scalar parameters and give no spatial
instruction on how the probe is to be guided, so detection reliability rests
entirely on operator execution and is further compromised by sensor signal
delays. This study quantifies how the kinematics of the manual sniffer
trajectory affect detection reliability at the small-scale standard pipes and
fittings common in hydrogen systems, using modular electrolyzer assemblies as a
representative case. This near field and low flow regime has been largely
neglected by existing macroscopic dispersion research. Using a robotically
guided test bench to eliminate operator-induced variability, static
concentration fields and dynamic trajectory passes were acquired across
representative component geometries under standardized leak rates of forming gas
(5 vol\% hydrogen in nitrogen) and varying scanning velocities. The results
demonstrate that scanning velocity, combined with spatial probe orientation,
strongly dictates detectability. Conventional linear scanning trajectories
frequently fail to register leaks under dynamic conditions, resulting in severe
false negatives. Conversely, geometry-specific routing, such as circumferential
plunging paths encircling the sealing point, maintains a high safety margin
across all tested scenarios. From these observations, geometry-specific routing
rules and a reduction-factor model for dynamic signal loss are derived. The
findings indicate that current standard operating procedures, which lack strict
spatial definitions, constitute a tangible safety risk. To operationalize these
rules, a proof-of-concept software pipeline is presented, which generates
validated trajectories directly from 3D models, usable for visualization in
assistance systems.
\end{abstract}

 
\end{frontmatter}

\section{Introduction}
\label{sec:intro}
 
Hydrogen is indispensable to applications ranging from aerospace propulsion to
ammonia and fertilizer production, and its use in modular water electrolyzers
is expanding the number of fluid-carrying joints that must be verified as
leak-tight during manufacturing. Because molecular hydrogen has a wide
flammability range in air and an extremely low minimum ignition energy, even
microscopic leaks at pressurized fittings can lead to ignition or
explosion~\cite{Toepler2017, Yang2021}, so reliable leak detection is
indispensable to quality assurance and system integrity.
 
To meet this requirement, non-destructive leak testing by tracer gas sniffing
is an obligatory acceptance procedure, in which an operator manually guides a
sniffer probe along every relevant surface of the assembly~\cite{ISO22734,
Masuhr.2024}. A modular electrolyzer is a representative case, containing on the
order of several thousand compression fittings to be inspected, yet the same
challenge applies to any densely packed hydrogen system built from standard
tubing and fittings. The procedure is thus both safety-critical and
time-intensive. As detailed in Section~\ref{sec:sota}, however, the governing
standards prescribe only scalar quantities such as test-gas pressure, tracer
concentration, and a maximum scanning rate, and contain no spatial instruction
for routing the probe around complex geometries~\cite{ISO22734, DIN20485}. The
reliability of the entire inspection thus rests on the intuition and experience
of the individual operator.
 
This reliance is precarious. The operator can neither perceive the invisible
tracer gas plume nor intuitively compensate for device-specific signal delays.
A slow electrochemical sensor that is moved marginally too fast, or held at a
sub-optimal angle, can miss a localized concentration entirely. Whether current
manual protocols can reliably detect the small, narrow leaks characteristic of
creeping fatigue failures is therefore an open question, and precisely the
question this study addresses.
 
To exploit these findings, operators could be given explicit guidance in
following a validated trajectory. Digital assistance systems can provide such
spatial guidance: an augmented-reality headset, for instance, can visualize
probe poses along a predefined route and remove the ambiguity from manual
positioning. This presupposes trajectories that are systematically designed to
guarantee detection. No validated trajectory generation yet exists for
sniffer-based leak detection (Section~\ref{sec:sota}).
 
To bridge the gap between error-prone manual execution and the algorithmic
reliability required by such assistance systems, this study investigates how
the trajectory of a sniffer probe, namely its path geometry, orientation, and
velocity, governs whether a small leak is detected at the complex component
topologies of hydrogen systems. The core contributions of this paper are:
\begin{enumerate}
    \item \textbf{Experimental framework:} a reproducible, robotically guided
    test bench that isolates the kinematic variables of sniffer-based leak
    detection from operator variability.
    \item \textbf{Static detectability fields:} an empirical characterization
    of the measurable concentration fields around small leaks, covering
    realistic leak geometries, mounting orientations, and outflow directions
    at leak rates spanning buoyancy-dominated to transition-regime dispersion,
    resolved separately for two industrially relevant detectors.
    \item \textbf{Dynamic trajectory effects and routing rules:} quantification
    of how scanning velocity and position-coupled probe orientation degrade
    the registered signal, condensed into a validated reduction-factor model
    for the dynamic signal loss and geometry-specific routing rules.
    \item \textbf{Automated trajectory generation:} a proof-of-concept pipeline
    that operationalizes these rules and generates validated sniffer
    trajectories directly from structured 3D component data for integration
    into assistance systems.
\end{enumerate}
 
The remainder of this paper is organized as follows. Section~\ref{sec:sota}
reviews the normative framework and the related scientific literature and
delimits the research gap. Section~\ref{sec:methods} describes the robotic test
bench, the leak geometries, and the static and dynamic measurement protocols.
Section~\ref{sec_results} reports the static concentration fields and the
dynamic trajectory results. Section~\ref{sec_rules} derives the
geometry-specific trajectory rules and the kinematic reduction-factor model,
and Section~\ref{sec_pipeline} presents the automated trajectory-generation
pipeline. Sections~\ref{sec_discussion} and~\ref{sec_conclusion} discuss the
implications for hydrogen safety guidelines and conclude.

\
\section{State of the Art and Related Work}
\label{sec:sota}
 
This section establishes the dual gap that motivates the study: standards
mandate leak testing but specify no spatial procedure, and the scientific
literature characterizes either the large-scale far field or the placement of
\emph{stationary} sensors, leaving the interaction between the near-field gas
plume and a \emph{moving} sniffer probe at component scale unaddressed.
 
\subsection{Regulatory framework and the normative gap}
\label{sec:sota:standards}
 
Hydrogen leak testing is a mandatory acceptance step for electrolyzer
assemblies and is embedded in a layered framework of regulations and
standards. In Germany, the Industrial Safety Regulation (BetrSichV) requires
continuous verification of system tightness~\cite{BetrSichV2015}, while
internationally ISO~22734 governs hydrogen generators using water
electrolysis~\cite{ISO22734}, ASME~B31.12 covers hydrogen piping and
pipelines~\cite{ASMEB3112}, and ISO~19880-1 sets the general requirements for
gaseous-hydrogen fuelling stations~\cite{ISO19880}. The procedural basis of
the sniffing technique is laid down in EN~1779 and ISO~20485, which place it
among the tracer-gas methods: the component is pressurized with a tracer gas
and a sniffing detector is guided around it to localize leaks from the
escaping concentration~\cite{DeutschesInstitutfurNormunge.V..1999, DIN20485}.
 
Across this framework, the prescriptions are almost exclusively \emph{scalar}
or \emph{component}-oriented. ISO~20485 fixes the test pressure, tracer
concentration, calibrated reference leak, and a maximum probe-scanning
rate~\cite{DIN20485}; ISO~22734 defers the leak-test procedure to the detector
manufacturer's instructions~\cite{ISO22734}; ASME~B31.12 mandates leak testing
and non-destructive examination but leaves the detection method
open~\cite{ASMEB3112}; and ISO~19880-1 treats leak detection chiefly through
fixed-sensor coverage and hazardous-area classification under
IEC~60079~\cite{ISO19880, IEC60079}. Detector-qualification standards such as
ISO~26142 likewise specify sensor performance rather than how a probe is to be
moved~\cite{ISO26142}.
 
That this landscape is fragmented and incomplete is documented in the review
literature. Qanbar and Hong~\cite{Qanbar.2024} survey the codes governing
hydrogen leak detection and find that the requirements emphasize sensor
specifications and zoning rather than inspection execution, and a recent
assessment of leak-detection and monitoring frameworks for hydrogen pipelines
concludes that existing provisions, largely adapted from natural gas, still
lack hydrogen-specific guidance, with leak-survey practices yet to be revised
for hydrogen's distinct risk profile~\cite{Dinata2025}. Even the acceptance
threshold is not harmonized: already in 2005 Block~\cite{Block2005} noted that
no generally binding permissible hydrogen leakage rates exist and that they are
standardized only in a few cases, and Chen et~al.~\cite{Chen.2025} confirm that
allowable rates still differ markedly between the GB and ISO standards, leaving
the threshold to be set per system in practice~\cite{Masuhr.2024}.
 
What no document in this framework provides is a \emph{spatial} specification
of the manual inspection itself: there is no rule for the stand-off distance,
local probe orientation, path geometry, or dwell time, so the execution is
relegated entirely to operator intuition. This is the first component of the
research gap.
 
\subsection{Manual sniffing and its kinematic sensitivity}
\label{sec:sota:methods}
 
Among non-destructive leak detection methods, a key distinction separates
\emph{integral} tests, which return the summed leakage of an enclosure without
localizing it, from \emph{local} tests, which pinpoint the
defect~\cite{Block2005, Qanbar.2024}. Since in-process inspection must
determine \emph{where} a fitting leaks, only local methods apply. Optical
techniques such as background-oriented schlieren resolve density gradients but
require highly controlled environments~\cite{Sun.2023}, and acoustic-emission
methods localize the sound of an escaping jet but sense its origin rather than
the gas itself, and thus cannot map where the detectable gas
resides~\cite{Zhang.2021}. For the small leaks at the complex fittings that
dominate electrolyzer assemblies, manual tracer-gas sniffing is therefore the
method of choice: it measures the local concentration, is portable and
selective to hydrogen, and is the only affordable, widely available technique
with sufficient sensitivity for leaks below 100 ml/min~\cite{Arai.2021,
Masuhr.2024, Block2005}.
 
The detector spans a wide range. Electrochemical sensors are compact and
inexpensive but respond slowly, because the gas must diffuse through membranes
and electrolytes before reacting, and their baseline can
drift~\cite{Saunders.2025}; mass spectrometers offer faster response and lower
detection limits at higher cost~\cite{Chen.2018, Saunders.2025}. This
dichotomy directly governs how a moving probe interacts with a localized plume.
Two effects make the sniffing result depend on \emph{how} the probe is moved,
not only \emph{where} it passes: the outcome is governed by sensor
response time and detection limit, and the probe is not a
passive observer, since its aspiration flow disturbs the local field, so that a
reliable reading requires quasi-static conditions while a probe moving too
quickly distorts the very plume it samples~\cite{GroeBley.2016}. The same slow,
exhaustive sampling makes static mapping impractically time-consuming, which
motivates dynamic scanning in practice. Yet no study relates
probe kinematics to detection reliability at the scale of individual fittings.
This is the second component of the research gap.
 
\subsection{Near-field dispersion and the role of leak geometry}
\label{sec:sota:dispersion}
 
Hydrogen dispersion has been studied extensively, but almost exclusively at
large scale and in the far field, in confined-chamber releases, refuelling
stations, and enclosed-space accumulation aimed at flammable-cloud and
explosion assessment~\cite{Lacome.2011, Xiao.2024, Yang2021}. These length
scales are irrelevant to locating a leak on a specific fitting. The near-field
regime at the leak itself is well described in principle: the densimetric
Froude number separates a buoyancy-dominated plume that rises almost
immediately at low release velocity from a momentum-dominated jet at high
velocity~\cite{Yang2021}, and at the low rates and pressures relevant here the
flow can remain laminar, with a quadratic dependence of leakage rate on
pressure~\cite{Schmid.2019, Block2005}.
 
A second limitation is the idealization of the leak as a perfect circular
hole~\cite{Lacome.2011}, whereas the realistic failure morphologies that
dominate fittings, valves, and flanges, namely circumferential gaps and
embrittlement cracks, are under-represented~\cite{Fischer.2024}. The studies
that do consider non-circular geometries consistently show that morphology
matters: Kang et~al.~\cite{Kang.2024} report higher initial momentum and faster
spread for rectangular than circular orifices; Qin et~al.~\cite{Qin.2024} find
only marginal mass-flow differences between circular and square holes of equal
area and show that fittings of different sizes leak in consistent patterns, a
result adopted in the present design (Section~\ref{sec:methods}) to justify a
single representative fitting; and Li et~al.~\cite{Li.2025} demonstrate
experimentally that circular, slit, and Y-type orifices produce markedly
different concentration fields, response times, and hazard rankings.
Critically, however, these geometry studies were conducted at elevated pressure
and resolved the field only in the far field (sensor stand-off of 0.25 m to
1 m); none resolves the near-field \emph{detectable} field around a complex
fitting at the close-range, low-pressure conditions of a manual sniffing
inspection.
 
\subsection{Detector-path planning and the research gap}
\label{sec:sota:planning}
 
A distinct body of work plans detector layouts from dispersion data. Sui
et~al.~\cite{Sui.2024} optimize the \emph{placement} of fixed sensors for
indoor leak detection, and Girotto et~al.~\cite{Girotto.2022} optimize the
\emph{trajectories} of gas detectors across a process plant by combining a
set-covering formulation with dispersion data. Both, however, operate at plant
scale, deciding where to mount stationary detectors or how to route a
facility-wide survey, not how to guide a hand-held probe over a single fitting,
and both treat the sensor as an ideal point detector, neglecting the
response-time lag and aspiration disturbance that dominate at close range. In
manufacturing metrology, the automated generation of inspection paths from CAD
data is established through Computer-Aided Inspection Planning
(CAIP)~\cite{zhaoComputerAidedInspectionPlanning2009a}, but CAIP toolpaths
follow solid surfaces; a sniffer samples a dynamic, invisible cloud, so these
algorithms cannot be transferred directly.
 
In summary, every reviewed standard constrains only scalar quantities and
delegates the spatial execution to intuition; large-scale dispersion studies
resolve the far field but not the near-field detectable plume; geometry studies
establish that morphology matters but stop at far-field, high-pressure
characterization; and detector-planning studies optimize stationary placement
while ignoring the kinematics of a moving probe. No standard or scientific work
therefore provides a validated basis for routing a manual sniffer probe around
the small, complex fittings of an electrolyzer while accounting for the coupled
near-field fluid dynamics and sensor kinematics, which is the objective of this
study.

\section{Methods and Experimental Setup}
\label{sec:methods}
 
\subsection{Objective and approach}
\label{sec:methods:objective}
 
The aim of the experimental campaign is to determine how the trajectory of a
sniffer probe governs whether a leak is detected. This is addressed in two
complementary steps. First, static concentration mapping establishes where a
measurable concentration exists around a component under stabilized conditions,
that is, the best case a trajectory could exploit. Second, dynamic trajectory
testing quantifies how the moving probe, subject to its finite response time,
degrades that signal. Linking the two yields the geometry-specific routing rules
of Section~\ref{sec_rules}. To acquire this data free from the tremor,
inconsistent stand-off, and varying velocity of a human operator, all
measurements were performed on a robotically guided test bench, so that the
recorded signal reflects the interaction of the gas plume with the sensor rather
than operator variability.
 
\subsection{Investigated components and leak geometries}
\label{sec:methods:geometries}
 
The study focuses on the components that dominate hydrogen assemblies by
frequency. In a representative modular electrolyzer pipes and fittings account for the overwhelming majority of sealing points to be inspected. Standard 1/4 inch pipe and fittings are prevalent across hydrogen infrastructure, therefore serving as the reference geometries. They are studied in horizontal and vertical mounting, as right anlge mounted components are industry standard which represents the majority of mounting orientations. Restricting the study to these recurring components keeps the findings transferable to any densely packed hydrogen system built from standard tubing and fittings.
 
Industrial leak morphologies were abstracted into three representative
geometries implemented on a reference pipe\cite{Li.2025}: a circular hole of 1 mm diameter representing a localized perforation (see Figure~\ref{fig:experiment_setup}). A slit of $1\times20$ mm representing the crack that develops through hydrogen embrittlement. A circumferential gap formed by assembling a compression fitting without its ferrules, representing the tolerance dependent sealing-point leakage typical of valves and fittings. The single gap dimension is representative, as varying the loosening angle and fitting size changes the flow rate only marginally~\cite{Qin.2024}. Because the outflow direction of a real defect is unknown a priori, the hole and slit were additionally investigated at outflow angles of 0, 90, and 180 degrees relative to the upward vertical, so that a trajectory can later be required to detect a leak independently of its orientation. In symmetrical experiment configurations we neglect the mesurement of an additional leak in 270° position. The chosen configurations repesent a limited but relevant basic test cases usbale for a time consuming test setup which enables the results to be transferable to other test cases.
 
\subsection{Operating conditions and gas supply}
\label{sec:methods:gas}

No universal threshold exists for an acceptable hydrogen leakage rate, as it
depends on the safety concept and risk-mitigation strategy of the individual
system~\cite{Masuhr.2024}. Three leak rates were therefore selected to reflect
both industrial requirements and the underlying gas behaviour. The rate of
100 mln/min represents the critical industrial threshold above which leaks must
be reliably detected. The lower rates of 50 and 20 mln/min were chosen to remain
within the measurement capability of both detectors while capturing a transition
in dispersion behaviour across the investigated leak geometries. This transition
is quantified by the densimetric Froude number, the ratio of momentum to
buoyancy force~\cite{Yang2021},
\begin{equation}
\label{eq:methods:froude}
Fr = \frac{U_{exit}}{\sqrt{g\,d_{noz}\,(\rho_{\infty}-\rho_{exit})/\rho_{exit}}},
\end{equation}
where $U_{exit}$ is the exit velocity, $g$ the gravitational acceleration,
$d_{noz}$ the characteristic orifice dimension (the hole diameter, and the 20 mm
slit length for the crack), and $\rho_{\infty}$ and $\rho_{exit}$ the ambient and
exit densities. Evaluating $Fr$ for the three rates (Table~\ref{tab:methods:froude})
shows that increasing the leak rate shifts the hole from buoyancy-influenced flow
toward a momentum-dominated jet, whereas the slit remains buoyancy-dominated throughout, so the two geometries probe qualitatively different dispersion regimes. As the Froude
number scales with the exit velocity, the same regimes are reproduced by a
smaller orifice at a higher rate, so the findings transfer to smaller leaks at
correspondingly higher flows; smaller leaks than those studied here quickly
fall below both industrial relevance and the detection limit of the
instruments. The fitting gemoetry is excluded, as its flow divides between two sealing points in an unknown ratio, precluding a meaningful exit velocity.

The test medium was forming gas (95 vol\% nitrogen, 5 vol\% hydrogen), a
non-flammable, cost-effective tracer widely used in industrial inspection. The
leak rate was set by two Bronkhorst FG-201-CV mass-flow controllers spanning
0 to 50 and 20 to 2000 mln/min, each calibrated for a distinct control interval,
and after each change a 120 s settling time purged the pipe. Here mln/min
denotes standard millilitres per minute (equivalent to sccm), and as the tracer
contains 5 vol\% hydrogen, the actual hydrogen flow is only 5\% of the regulated
rate.
 
\begin{table}[h]
\centering
\caption{Modified free-jet Froude number for the hole and slit geometries,
estimated from the exit velocity for the three leak rates. The hole lies in the
transition regime, whereas the slit remains buoyancy-dominated.}
\label{tab:methods:froude}
\setlength{\tabcolsep}{6pt}
\renewcommand{\arraystretch}{1.2}
\begin{tabular}{|l|c|c|c|c|}
\hline
\textbf{Geometry} & \textbf{Rate} & \textbf{Area $A$} & \textbf{$U_{exit}$} & \textbf{$Fr$} \\
                  & \textbf{(mln/min)} & \textbf{(m$^2$)} & \textbf{(m/s)} & \textbf{(-)} \\
\hline\hline
\multirow{3}{*}{Hole} & 20  & $7.85\cdot10^{-7}$ & 0.42   & 26.6  \\
                      & 50  & $7.85\cdot10^{-7}$ & 1.06   & 67.1  \\
                      & 100 & $7.85\cdot10^{-7}$ & 2.12   & 134.2 \\
\hline
\multirow{3}{*}{Slit} & 20  & $2.00\cdot10^{-5}$ & 0.0167 & 1.06  \\
                      & 50  & $2.00\cdot10^{-5}$ & 0.0417 & 2.64  \\
                      & 100 & $2.00\cdot10^{-5}$ & 0.0833 & 5.28  \\
\hline
\end{tabular}
\end{table}
 
\subsection{Sensor systems}
\label{sec:methods:sensors}
 
Two detectors spanning the industrial spectrum were evaluated
(Table~\ref{tab:sensor_specs}): a mid-tier mobile electrochemical instrument
(Dräger X-am 8000 with the XXS-H2 cell), with a high aspirated flow of about
500 sccm and a slow response~\cite{Draeger2023}, and a high-precision mass
spectrometer (Pfeiffer ASM 310) in sniffing mode, with a low aspirated flow of
about 60 sccm and a near-instantaneous response~\cite{Pfeiffer2024}. A high
aspirated flow widens the detectable region but dilutes the sample, while a slow response makes the registered signal lag the moving probe.

The dynamic behaviour of each detector was characterized in-house rather than
taken from the datasheet. Following a step-response procedure, the probe was
moved rapidly to a fixed 1 mm upward leak, held until the reading stabilized,
and withdrawn into hydrogen-free air, while pose and signal were recorded on a
common time base. From each step, the signal delay $D$ (the transit time through the sampling line) and the time constant $T_{90}$ were extracted, yielding
$T_{90}=7.9$ s with $D=9.4$ s for the XXS-H2 cell and $T_{90}<1$ s with
$D\approx1.1$ s for the mass spectrometer. The large delay and time constant of
the electrochemical cell are the origin of the dynamic signal loss modelled in
Section~\ref{sec_rules}.
 
\begin{table*}[t]
\centering
\caption{Specifications of the two tracer-gas detectors, illustrating the
contrast in aspiration mechanics and dynamic response.}
\label{tab:sensor_specs}
\setlength{\tabcolsep}{6pt}
\renewcommand{\arraystretch}{1.2}
\begin{tabular}{|l|c|c|}
\hline
\textbf{Parameter} & \textbf{Electrochemical} & \textbf{Mass spectrometer} \\
                   & \textbf{(Dräger X-am 8000, XXS-H2)} & \textbf{(Pfeiffer ASM 310)} \\
\hline\hline
Aspirated flow $Q$            & $\approx 500$ sccm        & $\approx 60$ sccm \\
\hline
Response time $T_{90}$ (measured) & $7.9$ s               & $<1$ s \\
\hline
Signal delay $D$ (measured)   & $9.4$ s                   & $\approx 1.1$ s \\
\hline
Detection method              & Electrochemical cell      & Quadrupole mass spec. \\
\hline
Primary output                & $H_2$ conc. (ppm) & Leak rate (mbar\,l/s) \\
\hline
Detection threshold (this study) & 100 ppm                & 50 ppm \\
\hline
Role                          & Mid-tier mobile device & High-precision reference \\
\hline
\end{tabular}
\end{table*}
 
\subsection{Robotic test bench}
\label{sec:methods:bench}
 
\begin{figure*}[!]
\centering
\includegraphics[width=\textwidth]{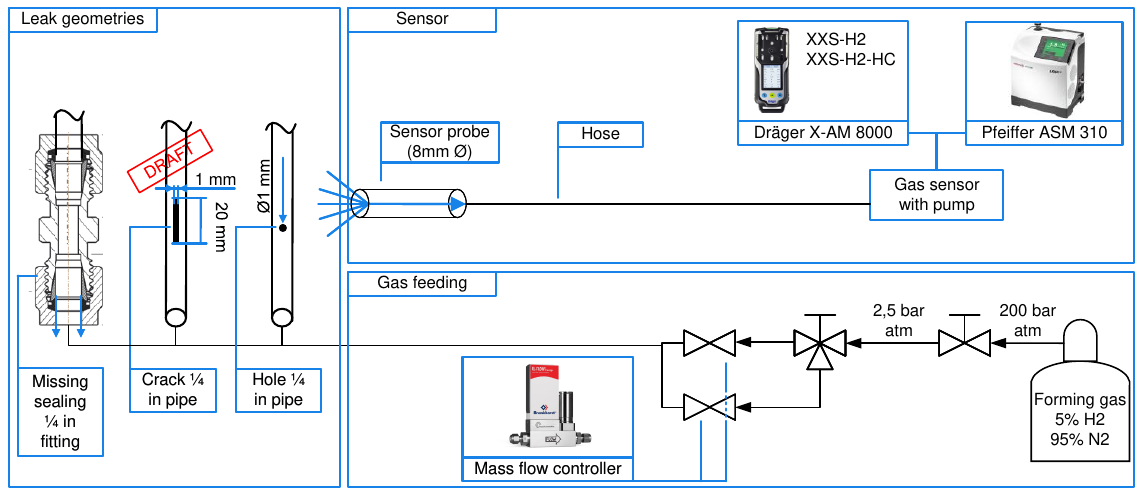}
\caption{Setup of experiment: Architectural overview of leak geometries, sensors and gas feeding used in the experiment.}
\label{fig:experiment_setup}
\end{figure*}

\begin{figure}[!t]
\centering
\includegraphics[width=\linewidth]{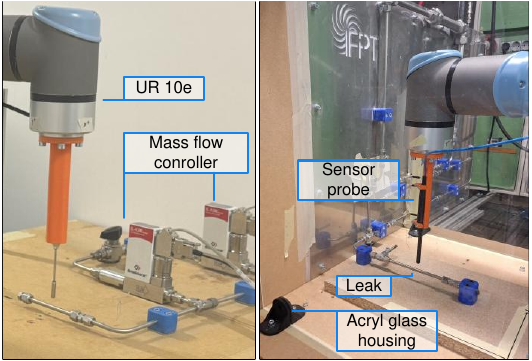}
\caption{Overview of the test setup: Robot with mounted pipes and plexiglass enclosing.}
\label{fig:experiment_subsystems}
\end{figure}

The probe was mounted as the tool center point of a collaborative six-axis
manipulator (Universal Robots UR10e, pose repeatability $\pm0.05$ mm) which conducted automated workflow controlled via a Robot Operating System (ROS2) application coordinating the manipulator and the flow
controllers. The bench was enclosed with Plexiglas on the accessible sides to shield
aerodynamic disturbances (Figure~\ref{fig:experiment_setup}), with the top left open to prevent accumulation of the buoyant tracer. The probe outer diameter was 6 mm (Figure~\ref{fig:experiment_subsystems}). The two instruments were measured in sequential
campaigns separated by a bench rebuild, each calibrated to the ambient background beforehand.
 
\subsection{Static and dynamic measurement protocols}
\label{sec:methods:protocols}
 
The procedure separates static concentration mapping from dynamic trajectory
testing, isolating the fluid-dynamic dispersion of the gas from the kinematic
interaction of the moving probe.
 
\begin{figure*}[!]
\centering
\includegraphics[width=\linewidth]{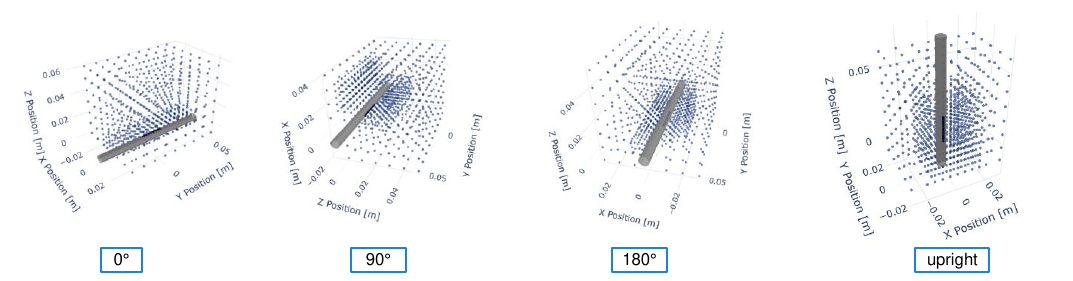}
\caption{Isometric view of the static measurement grid around the crack leak with three different leak angles, as well as an upright leak. Nodes are spaced 1 cm, refined to 0.5 cm in the near field, with unreachable nodes around the pipe removed.}
\label{fig:methods:raster}
\end{figure*}

\paragraph{Static concentration mapping}
The manipulator positioned the probe at the nodes of a three-dimensional grid
around the leak (Figure~\ref{fig:methods:raster}), with a nominal spacing of
1 cm refined to 0.5 cm in the near field of the electrochemical campaign, suited to the respective components and leak direction. At each node
a settling time (20 s electrochemical, 5 s mass spectrometer) excluded the
dynamic response, after which the signal was averaged over 10 s at 100 Hz. The
resulting node represents a trade of experiment time and resolution. Static
mapping focused on the critical boundary cases, and the concentration fields
shown in Section~\ref{sec_results} are reconstructed from these grids by cubic
interpolation. Thresholds were set per instrument: 100 ppm for electrochemical sensor, reliably detectable in pre-tests, and 50 ppm for the mass spectrometer sensor. A trajectory detects a leak only where a
concentration above the threshold is guaranteed regardless of geometry and
outflow direction.

\paragraph{Dynamic trajectory testing}
The bench executed continuous paths at 10, 20, and 30 mm/s, with the probe
orientation varied from orthogonal to grazing, each combination repeated five
times. Pose and signal were recorded per pass, the maximum concentration taken
as the detection indicator (a pass counts as a detection when it exceeds the
threshold), and each path was extended by start and end segments outside the leak
influence. These tests quantify the spatial signal drag from the response time
and the degradation under sub-optimal orientation and velocity.
 
\paragraph{Design of experiments}
Table~\ref{tab:doe} summarizes the resulting design: the three geometries, the
relevant pipe orientations and outflow directions, and the three leak rates, each
mapped statically with both detectors and traversed dynamically at the three
velocities with five repetitions. Supplementary cases (a low-velocity background
draft and a nearby boundary wall) probe the robustness of the findings.
 
\begin{table*}[t]
\centering
\caption{Design of static experiments parameters conducted with both
detectors.}
\label{tab:doe}
\setlength{\tabcolsep}{6pt}
\renewcommand{\arraystretch}{1.2}
\begin{tabular}{|c|l|c|c|c|}
\hline
\textbf{Case} & \textbf{Geometry} & \textbf{Pipe orient.} & \textbf{Outflow direction} & \textbf{Leak rate (mln/min)} \\
\hline\hline
1 & Hole (1 mm)            & Horizontal & 0/90/180 deg        & 20, 50, 100 \\
2 & Hole (1 mm)            & Vertical   & axial (symmetric)   & 20, 50, 100 \\
3 & Slit ($1\times20$ mm)  & Horizontal & 0/90/180 deg        & 20, 50, 100 \\
4 & Slit ($1\times20$ mm)  & Vertical   & axial (symmetric)   & 20, 50, 100 \\
5 & Fitting (0.1 mm gap)   & Horizontal & radial (symmetric)  & 20, 50, 100 \\
6 & Fitting (0.1 mm gap)   & Vertical   & radial (symmetric)  & 20, 50, 100 \\
\hline
\end{tabular}
\end{table*}
 
\newpage

\section{Experimental Results}
\label{sec_results}
 
The experimental campaign yielded static concentration maps and dynamic
trajectory passes across the geometries, orientations, and leak rates of
Section~\ref{sec:methods}. Static maps (Section~\ref{sec:res:static}) establish
where a measurable concentration exists under stabilized conditions, and
dynamic passes (Section~\ref{sec:res:dynamic}) show how the moving probe
degrades it; the quantitative evaluation that selects the preferred path per
topology follows in Section~\ref{sec_rules}.
 
\subsection{Static concentration mapping}
\label{sec:res:static}
 
The static maps represent the maximum measurable signal at each point and thus
describe the best case a trajectory could exploit, before efficiency and the
dynamic sensor behaviour are considered.
 
\subsubsection{Influence of aspiration flow}
\label{sec:res:static:aspiration}
 
\begin{figure}[!t]
\centering
\includegraphics[width=\linewidth]{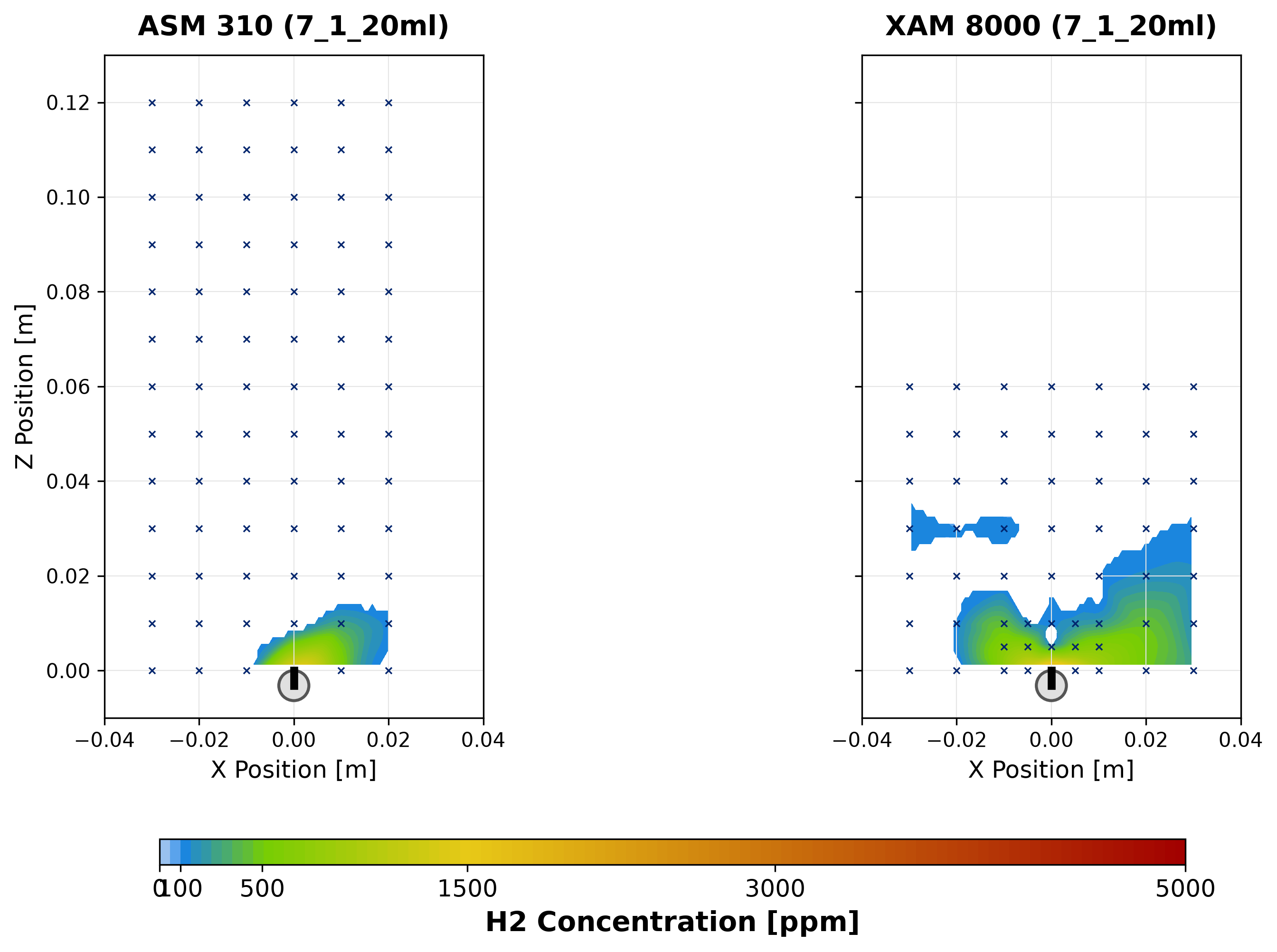}
\caption{Static concentration fields at a slit leak (20 mln/min). The confined
zone of the mass spectrometer (left) contrasts with the broader detectable
field of the high-aspiration electrochemical sensor (right). Iso-lines mark
the detection thresholds.}
\label{fig:res:static_aspiration}
\end{figure}

The static fields are detector-specific. The high aspirated flow of the
electrochemical sensor actively draws surrounding gas toward the probe and
thereby widens the region in which the leak is registered, whereas the
low-aspiration mass spectrometer samples an almost undisturbed field and
detects the tracer only in a confined zone near the leak
(Figure~\ref{fig:res:static_aspiration}). The maps therefore represent
detectable fields rather than pure concentration distributions, an effect that
applies equally during real inspection, where the probe's intake perturbs the
plume it samples~\cite{GroeBley.2016}. The contrast between the two instruments
is sharpest at the lowest leak rate.
 
\subsubsection{Geometry, outflow direction, and leak-rate regime}
\label{sec:res:static:orientation}
 
\begin{figure*}[!t]
\centering
\includegraphics[width=\linewidth]{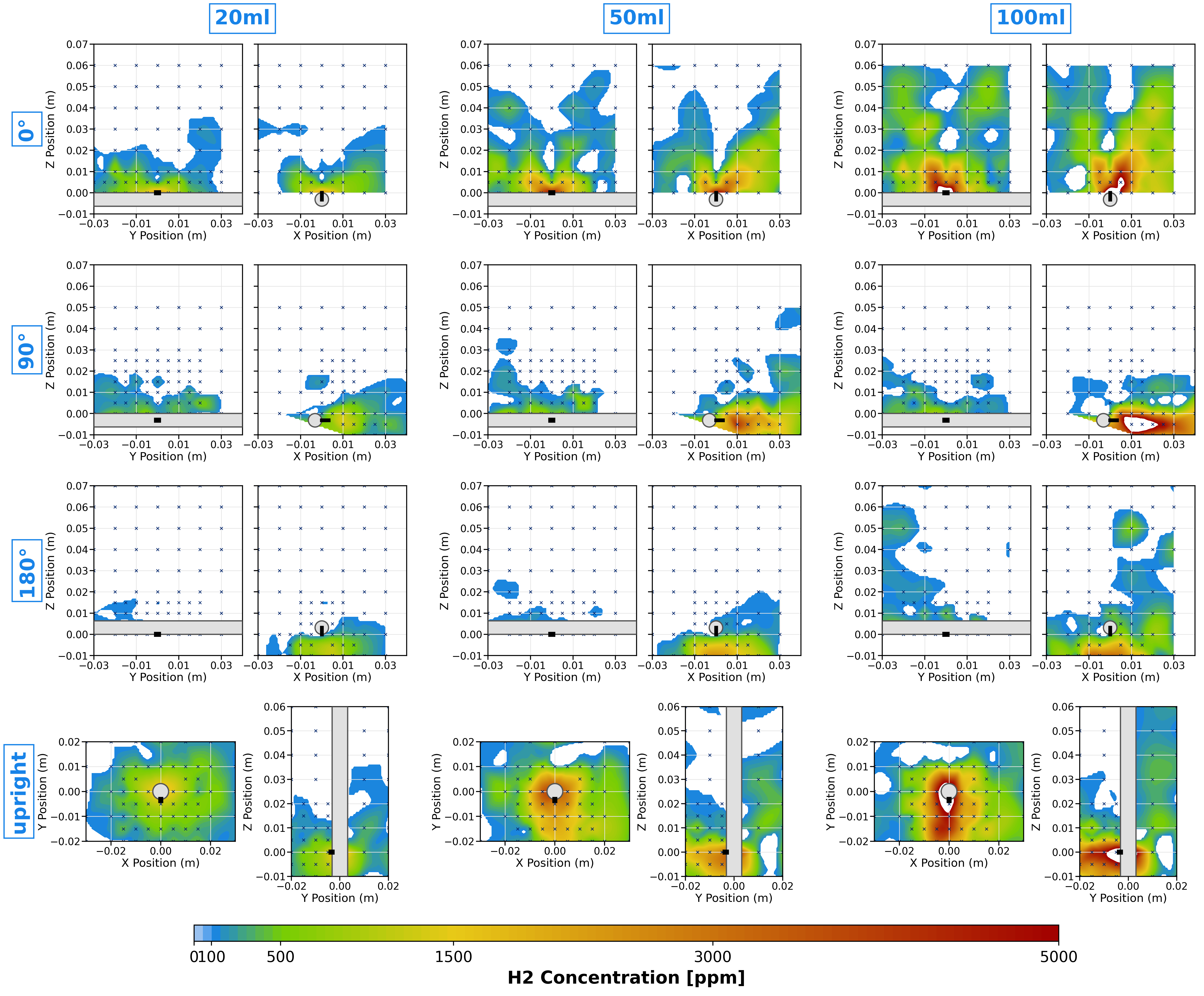}
\caption{Interpolated static concentration fields at the hole (XXS-H2 cell),
for outflow directions of 0, 90, and 180 degrees in horizontal mounting and
for vertical (upright) mounting (rows), at leak rates of 20, 50, and
100 mln/min (columns), each shown from two orthogonal viewing planes;
measurement nodes are marked and the pipe is shaded grey. White color represents measurement values below the sensor threshold of 100 ppm H2.}
\label{fig:res:static_hole}
\end{figure*}

\begin{figure*}[!t]
\centering
\includegraphics[width=\linewidth]{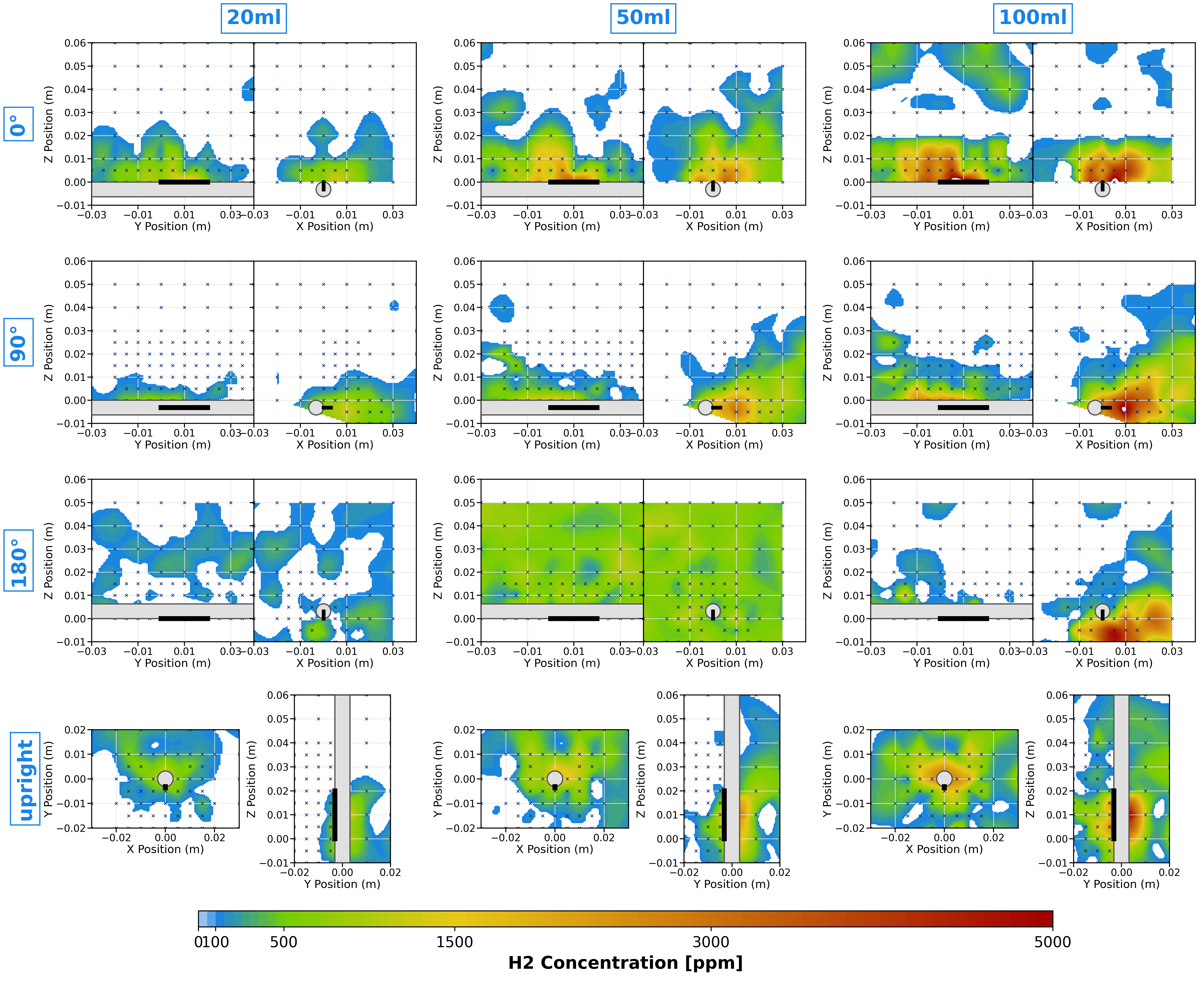}
\caption{Interpolated static concentration fields at the slit (XXS-H2 cell),
for outflow directions of 0, 90, and 180 degrees in horizontal mounting and for
vertical (upright) mounting (rows), at leak rates of 20, 50, and 100 mln/min
(columns), each shown from two orthogonal viewing planes. Coverage above the
pipe remains broad when the slit points upward or in vertical mounting, but
narrows markedly when it points sideways or downward. White color represents measurement values below the sensor threshold of 100 ppm H2.}
\label{fig:res:static_slit}
\end{figure*}

The mounting orientation reshapes the dispersion: a horizontal pipe lets the
gas rise to a maximum above the component, while a vertical pipe makes it cling
to and rise along the surface. The unknown outflow direction is decisive. For a
horizontal pipe with the hole pointing sideways (90 degrees), the gas forms a
lateral jet that rises only several centimetres downstream, shifting the
detectable field off the top of the pipe (Figure~\ref{fig:res:static_hole});
the slit shows a comparable direction dependence
(Figure~\ref{fig:res:static_slit}). The upward zenith is nevertheless the only position intersecting a detectable concentration for all
tested outflow directions at once, which underlies the routing rule of
Section~\ref{sec_rules}. The mass spectrometer produces the more critical
fields: at 20 mln/min the maximum stand-off at the slit, limited by its
sideways outflow, was only about 0.25 cm, rising to 0.5 cm at 50 mln/min, and
at 100 mln/min two critical cases emerged, with no measurable region left above
the pipe for the sideways or downward pointing hole, and even at the slit's
least favourable cross-section along its length, only a minimal, side-dependent
region remained when pointing downward.
 
The leak rate itself changes the dispersion regime, as anticipated by the
Froude analysis (Equation~\ref{eq:methods:froude},
Table~\ref{tab:methods:froude}). At the vertical pipe, the hole at 20 mln/min
produces a uniform, cloud-like spread with large measurable areas even on the
far side of the pipe, whereas at 100 mln/min the gas concentrates into a
directed jet: far-side spread decreases, and the readings directly above the
hole stagnate or even fall despite the fivefold rate
(Figure~\ref{fig:res:static_hole}). The hole combined with a high leak rate is
therefore the critical case for trajectory placement, with the significant
region shrinking to a height of partly below 2 cm and only low readings
immediately at the pipe on the far side. The slit, remaining buoyancy-dominated
at all rates, keeps an even distribution around the pipe with a significant
region partly exceeding 4 cm, its radial spread growing from about 1 cm at
20 mln/min to 2.5 to 3 cm at 100 mln/min (Figure~\ref{fig:res:static_slit}).
 
\begin{figure*}[!t]
\centering
\includegraphics[width=\linewidth]{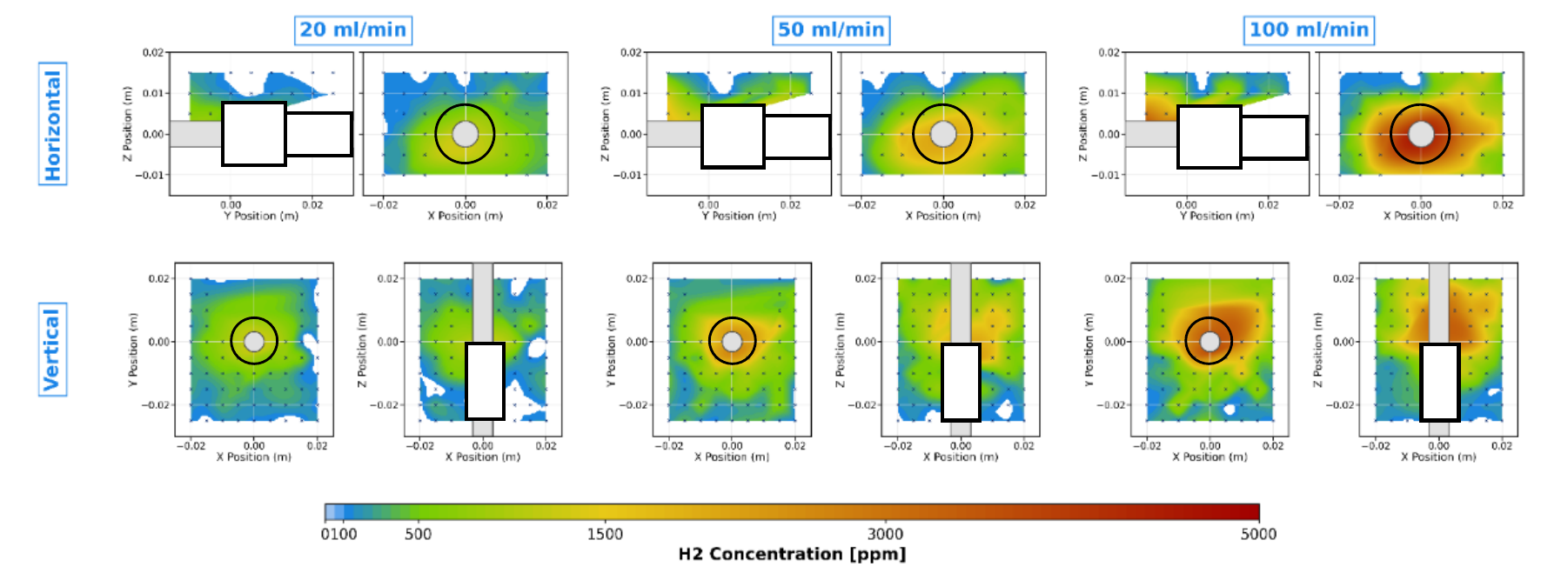}
\caption{Interpolated static concentration fields at the compression fitting
(XXS-H2 cell), in horizontal and vertical mounting (rows) at leak rates of 20,
50, and 100 mln/min (columns), each shown from two orthogonal viewing planes;
measurement nodes are marked and the fitting is shaded grey. High readings
concentrate around the union nut and intensify with the leak rate. White color represents measurement values below the sensor threshold of 100 ppm H2.}
\label{fig:res:static_fitting}
\end{figure*}

At the fittings, the gas escapes at both ends of the union nut, with the
dominant share at the transition from fitting to pipe, and very high readings
occur radially within 1.5 to 2 cm of the sealing points
(Figure~\ref{fig:res:static_fitting}). Despite the rotationally symmetric
geometry, the outflow is distributed unevenly around the circumference. In
horizontal mounting this has little consequence, as a layer of high readings
forms above the fitting along nearly its whole length; in vertical mounting it
produces a distinctly uneven distribution around the nut, the basis for the
vertical-fitting rule of Section~\ref{sec_rules}.
 
\subsubsection{Orientation sensitivity and its coupling to position}
\label{sec:res:static:angle}
 
\begin{figure}[!t]
\centering
\includegraphics[width=\linewidth]{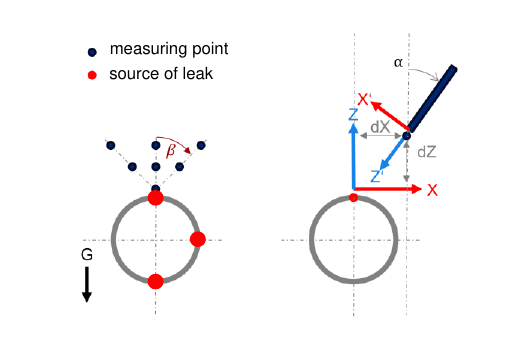}
\caption{Definition of the position-coupled orientation test. The probe tilt
angle $\alpha$ and circumferential test angle $\beta$ are measured relative to
the component axis, and the lateral and axial displacement ($dX$, $dZ$)
between the probe tip and the leak source defines the tested positions around
the component circumference.}
\label{fig:res:configurations}
\end{figure}

\begin{figure*}[!t]
\centering
\includegraphics[width=\linewidth]{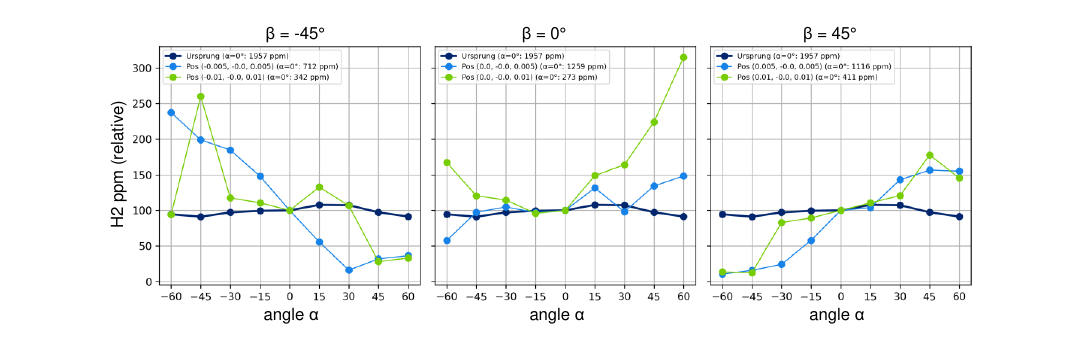}
\caption{Measured H2 concentration, relative to the centred, surface-normal
reference (Ursprung, $\alpha=0^\circ$), as a function of probe rotation angle
$\alpha$, for three circumferential test positions $\beta$ and several lateral
and axial displacements from the leak ($dX$, $dZ$; legend values give the
concentration in ppm at $\alpha=0^\circ$ for each displaced position). }
\label{fig:res:static_angle}
\end{figure*}

Manual execution varies both probe angle and position, examined 
separately in the experiment and jointly using the position-coupled orientation test defined in
Figure~\ref{fig:res:configurations} (probe tilt angle $\alpha$, circumferential
test angle $\beta$, and lateral or axial displacement $dX$, $dZ$ between the
probe tip and the leak source). With the probe centred
above the leak and rotated about its tip, the reading was robust: for the mass spectrometer it stayed essentially constant up to 50 to 60 degrees, and across instruments a reduction appeared only beyond
about 75 degrees (roughly 10 percent at 75 degrees, up to 25 percent at 90
degrees). When orientation is coupled to position, however, it becomes decisive:
with the probe displaced diagonally from the leak and rotated at each position,
a probe displaced to one side and angled away from the pipe axis lost 70 to 90
percent of the signal.Once the tip is carried beyond the pipe axis the
signal collapses regardless of angle (Figure~\ref{fig:res:static_angle}). The
dominant failure mode is therefore not a small tilt but lateral displacement combined with a probe no longer
pointing at the surface: detectability is preserved as long as the probe points
at the surface and its tip does not pass the component axis, an observation
formalized as a routing constraint in Section~\ref{sec_rules}.

The static maps thus give a deterministic baseline: the detectable gas is
dictated by topology, outflow direction, aspiration, and the leak-rate regime.
A single straight line is not universally valid and the trajectory must adapt
to the geometry to intersect a detectable concentration regardless of the
unknown defect orientation.
 
\subsection{Dynamic trajectory interaction}
\label{sec:res:dynamic}
 
While the static maps show where gas is measurable in principle, the dynamic
passes reveal what a moving probe registers. Given the combinatorial scope of
geometry, orientation, outflow direction, rate, and velocity, the dynamic
protocol focused on selected, critical configurations identified from the
static maps (Section~\ref{sec:methods:protocols}). Because the trajectories
target the AR assistance system built around the electrochemical instrument,
these passes used primarily the XXS-H2 cell; the mass spectrometer was
characterized dynamically for a single additional configuration, discussed at
the end of this section.
 
\subsubsection{Spatial signal drag and velocity limits}
\label{sec:res:dynamic:drag}
 
The slow electrochemical response (measured $T_{90}=7.9$ s) causes a spatial
signal drag: crossing a localized plume at constant velocity, the registered
peak lags the leak and its amplitude falls. A first-order sensor model
parameterized by the measured response time reproduces both effects, and the
loss is severe at sniffing speeds; the model and its validation against the
measured passes are developed in Section~\ref{sec:rules:model}. This explains
why a fast scan with a slow sensor produces false negatives at small leaks.
 
\subsubsection{Velocity dependence of detection at a vertical pipe}
\label{sec:res:dynamic:vertical}
 
\begin{figure*}[!t]
\centering
\includegraphics[width=\linewidth]{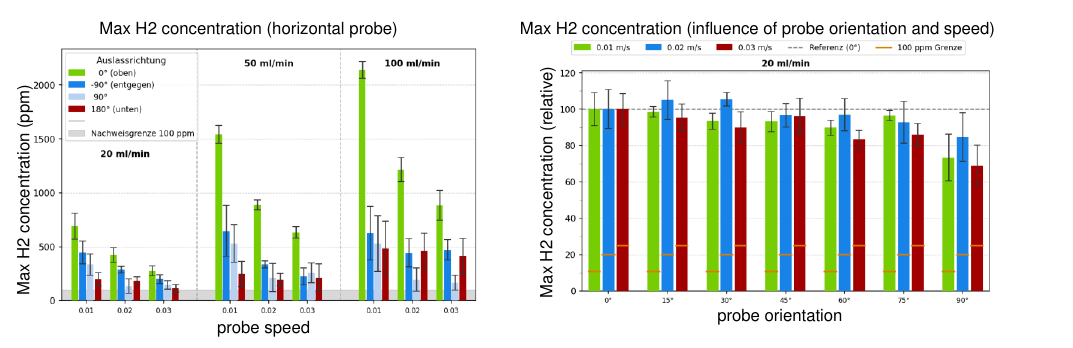}
\caption{Maximum registered concentration for the straight-line scan along a
vertical pipe, by velocity and leak rate (five repetitions each); the dashed
line marks the 100 ppm threshold, and detection drops to 80 percent at
20 mm/s (100 mln/min) and to 60 percent at 30 mm/s (20 mln/min) (left).
Maximum registered concentration, relative to the surface-normal reference,
for a straight-line scan across an upward-pointing hole as the probe
orientation is rotated, shown for three scanning velocities (right).}
\label{fig:res:vertical_straight}
\end{figure*}

The clearest demonstration of the safety risk is the straight-line scan along a
vertical pipe, with the leak on the far side of the pipe, facing away from the
scan line, as the critical case. At 10 mm/s every repetition exceeded the
100 ppm threshold at all leak rates. At 20 mm/s detection was reliable only at
20 and 50 mln/min, with some passes failing at 100 mln/min (80 percent
detection). At 30 mm/s the values fell so that at the lowest rate of 20 mln/min
only 60 percent of passes detected the leak
(Figure~\ref{fig:res:vertical_straight}). A straight line at a plausible manual
speed therefore misses the leak in a substantial fraction of attempts, for the given sensor.
 
By contrast, the laterally plunging path on the same pipe detected the leak in
every repetition with only a weak velocity dependence, the peak falling merely
from about 616 ppm at 10 mm/s to 537 ppm at 30 mm/s, because it repeatedly
enters the high-concentration region near the surface. The dynamic passes thus
prove that the static field does not equal the registered signal: velocity
coupled with the response time drives false negatives on straight paths, while
paths that enter the near-surface zone stay robust, so the temporal execution
matters as much as the spatial route.
 
\subsubsection{Probe orientation at a horizontal pipe}
\label{sec:res:dynamic:orientation}
 
A second dynamic validation targeted probe orientation itself, using the
horizontal-pipe hole across the same rate and velocity grid, with the outflow
direction (0, 90, and 180 degrees) as a further variable and the vertical and
horizontal probe orientations as the two bounding cases. The vertical probe,
pointing down onto the pipe from directly obove, detected every combination above the 100 ppm
threshold. The horizontal probe was reliable only at 10 mm/s: as velocity and
leak rate increased, the sideways outflow direction became critical, and at
100 mln/min the readings approached the detection limit closely enough that
reliable detection could no longer be assumed. Because the probe remains
centred on the pipe axis throughout, this is consistent with the
position-coupled orientation behaviour of Section~\ref{sec:res:static:angle}:
intermediate orientations remained detectable provided the velocity was reduced
toward the horizontal extreme. This is corroborated by a direct dynamic sweep
of probe orientation during a straight-line scan across an upward-pointing
hole (Figure~\ref{fig:res:vertical_straight}, right): the registered peak
stayed close to the centred reference across most of the tested angle range at
all three scanning velocities before falling off toward the largest rotation
angles, mirroring the static position-coupled orientation behaviour above.
Orientation is thus a secondary factor at a horizontal pipe compared with velocity, so long as the
probe stays centred on the pipe axis; the corresponding guidance is formalized in
Section~\ref{sec_rules}.
 
A dynamic check was additionally carried out with the mass spectrometer for the
upward pointing leak at 20 mln/min, sampled over a wide velocity range of 10 to
200 mm/s. The peak reading fluctuated between about 8200 and 13200 ppm with no
systematic decline, two orders of magnitude above the 50 ppm detection limit;
only the scatter across repeated passes grew markedly at higher velocities,
with the lower range of individual passes falling to about 4000 to 5000 ppm at
the highest speeds, still far above the threshold. Since the static maps
identify the electrochemical instrument as governing the narrower, more
critical detectable fields (Section~\ref{sec:res:static:orientation}), the
detailed dynamic characterization and the resulting rules focus on the XXS-H2
cell as the more velocity-sensitive instrument.

\section{Derivation of Geometry-Specific Trajectory Rules}
\label{sec_rules}
 
The experimental results show that the reliability of manual inspection is
compromised by uncontrolled kinematics. To move from intuition to explicit
constraints, this section first consolidates the dynamic passes into a
quantitative evaluation (Section~\ref{sec:res:eval}), then derives a
velocity-based signal-reduction model (Section~\ref{sec:rules:model}),
universal kinematic rules (Section~\ref{sec:rules:kinematic}), and
topology-specific routing rules (Section~\ref{sec:rules:topology}),
consolidated in Table~\ref{tab:rules:summary} as the database logic for the
automated pipeline of Section~\ref{sec_pipeline}. All quantitative values refer
to the electrochemical XXS-H2 cell, on which the dynamic characterization is
based (Section~\ref{sec:res:dynamic}).
 
\subsection{Quantitative trajectory evaluation}
\label{sec:res:eval}
 
The dynamic passes (Section~\ref{sec:res:dynamic}) showed that the registered
signal, and with it the margin against the detection limit, depends on
scanning velocity. To select a preferred trajectory per topology, each
candidate pass is therefore scored against this margin, together with its
execution effort, at the critical leak rate of 20 mln/min. The conservative
safety margin
\begin{equation}
\label{eq:res:sm}
SM_{kons} = \frac{\bar{H}_{2,\max} - k\,\sigma}{LOD}
\end{equation}
relates the mean peak concentration, reduced by the observed scatter ($k=1$,
justified by the small sample of five repetitions), to the detection limit
$LOD$: a value of $1$ just meets the threshold, higher values indicate
robustness. Execution effort is compared through a normalized time, noting that
plunging paths are about 13 percent longer on pipes and need 90 mm per sealing
point on fittings versus 20 mm for a straight pass, and path complexity is
rated qualitatively. Table~\ref{tab:res:eval} consolidates the evaluation. The
pattern is consistent across all four topologies: straight paths are
time-efficient but their margin collapses as velocity rises, whereas encircling
paths preserve a high, velocity-tolerant margin. The topology-specific
consequences are drawn in Section~\ref{sec:rules:topology}.
 
\begin{table*}[t]
\centering
\caption{Quantitative trajectory evaluation at the critical leak rate of
20~mln/min, shown for the slowest and fastest tested velocity. $\bar{H}_{2,\max}\pm\sigma$
is the mean peak over five repetitions; $SM_{kons}$ uses $k=1$ and the
electrochemical $LOD$ of 100~ppm. Values $SM_{kons}<1$ denote failed detection
(bold).}
\label{tab:res:eval}
\setlength{\tabcolsep}{6pt}
\renewcommand{\arraystretch}{1.15}
\begin{tabular}{|l|l|c|c|c|c|}
\hline
\textbf{Topology} & \textbf{Trajectory (probe)} & \textbf{$v$} & \textbf{$\bar{H}_{2,\max}\pm\sigma$} & \textbf{$SM_{kons}$} & \textbf{Compl.} \\
                  &                             & \textbf{(mm/s)} & \textbf{(ppm)} &        &  \\
\hline\hline
\multirow{4}{*}{Vertical pipe}
  & Straight (axial)   & 10 & $226\pm21$ & 2.0 & low \\
  & Straight (axial)   & 30 & $119\pm48$ & \textbf{0.7} & low \\
  & Plunging (axial)   & 10 & $616\pm24$ & 5.9 & med. \\
  & Plunging (axial)   & 30 & $537\pm52$ & 4.9 & med. \\
\hline
\multirow{3}{*}{Horizontal pipe}
  & Straight (horiz.)  & 10 & $198\pm62$ & 1.3 & low \\
  & Straight (vert.)   & 10 & $260\pm53$ & 2.1 & low \\
  & Straight (vert.)   & 30 & $188\pm53$ & 1.4 & low \\
\hline
\multirow{4}{*}{Horizontal fitting}
  & Straight (horiz.)  & 30 & $166\pm50$ & 1.1 & low \\
  & Straight (vert.)   & 30 & $213\pm29$ & 1.8 & low \\
  & Plunging (vert.)   & 10 & $498\pm72$ & 4.3 & med. \\
  & Plunging (vert.)   & 30 & $338\pm31$ & 3.1 & med. \\
\hline
\multirow{3}{*}{Vertical fitting}
  & Straight (axial)   & 10 & $257\pm57$ & 2.0 & low \\
  & Straight (axial)   & 30 & $146\pm46$ & 1.0 & low \\
  & Plunging (lateral) & 30 & $361\pm96$ & 2.7 & med. \\
\hline
\end{tabular}
\end{table*}
 
\subsection{Velocity-dependent signal-reduction model}
\label{sec:rules:model}
 
The dynamic passes (Section~\ref{sec:res:dynamic}) showed that a moving probe
registers only a fraction of the static peak concentration. To make this usable
as a design constraint, the dynamic peak is related to the static value through
a velocity-dependent reduction factor. Treating the sensor as a first-order
element with time constant $\tau$, the measured concentration follows
\begin{equation}
\label{eq:rules:ode}
\tau\,\frac{dC_{meas}}{dt} = C_{real}(t) - C_{meas}(t),
\qquad \tau = \frac{T_{90}}{2.3},
\end{equation}
where $C_{real}(t)$ is the concentration encountered along the path, obtained
by traversing the measured static field at velocity $v$. Integrating
Equation~\ref{eq:rules:ode} for a given path yields the dynamic peak
$C_{max}(v, T_{90})$, and the signal reduction is expressed as the factor
\begin{equation}
\label{eq:rules:Fv}
F_{v}(v, T_{90}) = \frac{C_{0}}{C_{max}(v, T_{90})} \ge 1,
\end{equation}
where $C_{0}$ is the static (stabilized) concentration. A factor of $F_v = 1$
means no loss, while larger values quantify how much of the true signal is lost
to the sensor lag during motion.
 
\begin{figure}[!t]
\centering
\includegraphics[width=\linewidth]{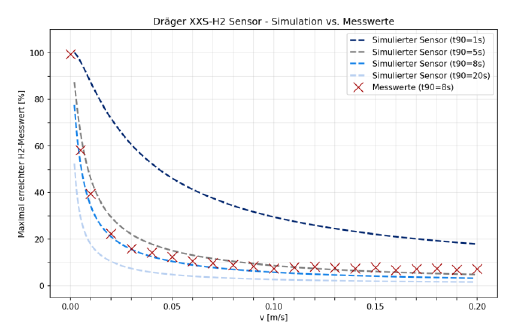}
\caption{Velocity-dependent signal reduction and model validation. Simulated
maximum sensor response when traversing a representative concentration field,
shown for several response times $T_{90}$, with the measured peaks of the
Dräger XXS-H2 cell ($T_{90}=7.9$ s) overlaid. Faster scanning and slower
sensors both reduce the registered peak, and the measured points track the
simulated 8 s curve, confirming the first-order model in the regime relevant to
manual inspection.}
\label{fig:rules:Fv}
\end{figure}

Evaluating Equation~\ref{eq:rules:Fv} for a representative plume (a 5 cm path
with a central peak, matching the upward hole and slit cases) gives the
behaviour in Figure~\ref{fig:rules:Fv}. The loss appears already at low speed:
at a typical sniffing speed of 20 mm/s, a fast sensor ($T_{90} = 1$ s) still
retains about 70 percent of the static value, whereas a slow sensor
($T_{90} = 8$ s) retains only about 21 percent. The model was validated against
the XXS-H2 cell, whose response time was measured as $T_{90} = 7.9$ s: over the
range 0 to 30 mm/s the measured peaks closely match the simulated
$T_{90} = 8$ s curve, confirming the model in the regime relevant to manual
inspection. A first check for a second instrument follows the same logic: with
a measured response time below 1 s for the mass spectrometer, the model
predicts negligible loss up to at least 200 mm/s, consistent with the absence
of a systematic decline observed for that instrument in
Section~\ref{sec:res:dynamic:orientation}.
 
Equation~\ref{eq:rules:Fv} allows a trajectory to be designed backward
from a safety requirement: given the sensor's measured $T_{90}$, the plume
extent along the path, and the margin required above the detection limit, it
yields the fastest velocity at which the dynamic peak still clears that
margin. Two caveats keep this a planning aid rather than a fixed number.
First, the factor is field-specific, not universal: a longer
high-concentration stretch, for example above a fitting, gives the signal more
time to develop and thus permits a higher velocity, so $F_v$ must be evaluated
for the field being traversed rather than looked up once. Second, it captures
the velocity effect only; the coupled effect of probe orientation is treated
separately in the rule below.
 
\subsection{Universal kinematic rules}
\label{sec:rules:kinematic}
 
Two constraints apply across all topologies.
 
\paragraph{Velocity coupled to sensor response}
To prevent false negatives, the maximum trajectory velocity must be tied to the
sensor response time and the minimum expected plume extent through
Equation~\ref{eq:rules:Fv}. For slow electrochemical sensors the velocity must
be restricted so that the probe dwells in the detectable field long enough for
the signal to develop. Where a low velocity is not acceptable, the trajectory
must instead insert short stationary dwell points at critical nodes to allow
the reading to stabilize.
 
\paragraph{Position-coupled orientation}
The orientation measurements (Section~\ref{sec:res:static:angle}) showed that a
centred probe is insensitive to tilt up to about 75 degrees, but that a probe
displaced from the leak and angled away from the surface loses 70 to 90 percent
of the signal. Orientation therefore cannot be treated by a simple angular law;
it matters only in combination with position. The resulting rule is geometric:
at every point of the path the probe must point toward the component surface
(surface-normal orientation), and its tip must never be carried past the
component axis. Between the surface-normal and horizontal extremes,
intermediate orientations remain admissible if the velocity is reduced
accordingly (Section~\ref{sec:res:dynamic:orientation}). When this rule is
respected, the orientation-related loss is negligible and the dynamic signal
reduction is governed by the velocity factor $F_v$ alone; the trajectory
generator enforces it by aligning the probe axis with the local surface normal
at each interpolation point.
 
\subsection{Topology-specific routing rules}
\label{sec:rules:topology}
 
The topology-specific rules follow from Table~\ref{tab:res:eval} and are
consolidated in Table~\ref{tab:rules:summary}. For a \emph{horizontal pipe},
buoyancy lifts the gas into a detectable band above the pipe that covers all
outflow directions, so a straight scan along the upper zenith suffices; a
vertically held probe is preferred, holding a margin above 2 at 10 mm/s and
still 1.4 at 30 mm/s, a horizontal probe is admissible only at 10 mm/s, and
intermediate orientations are usable with correspondingly reduced velocity. For
a \emph{vertical pipe} the straight axial line is not retained: its margin
falls from 2.0 at 10 mm/s to a failed 0.7 at 30 mm/s, whereas the laterally
plunging path holds 4.9 to 5.9 and, run at 30 mm/s, completes the pipe in less
total time despite its longer path. For a \emph{horizontal fitting}, buoyancy
is again exploited: a straight pass above the union nut with a vertical probe
is preferred, retaining a margin of 1.8 even at 30 mm/s, while the plunging
loop raises the margin to between 3.1 and 4.3 at roughly fivefold time per
sealing point and is reserved for cases the straight pass cannot cover. For a
\emph{vertical fitting} the straight line reaches only 1.0 at 30 mm/s and,
given the uneven circumferential outflow established in
Section~\ref{sec:res:static:orientation}, risks lying on a non-detecting side;
the plunging loop, holding 2.7 at 30 mm/s and inherently covering all radial
outflow directions, is therefore required.
 
\subsection{Synthesis of the rule set}
\label{sec:rules:synth}
 
The rules replace operator intuition with a deterministic framework that fixes,
per topology, the path geometry, probe orientation, and admissible velocity
(Table~\ref{tab:rules:summary}). The central trade-off is between a fast
straight path and a more complex plunging path that trades time for a much
larger, velocity-tolerant margin, and these rules serve as the database logic
for the pipeline of Section~\ref{sec_pipeline}.
 
\begin{table*}[t]
\centering
\caption{Geometry-specific trajectory rules for the electrochemical XXS-H2
sensor; safety margins $SM_{kons}$ at the critical leak rate of 20 mln/min from
Table~\ref{tab:res:eval}.}
\label{tab:rules:summary}
\setlength{\tabcolsep}{5pt}
\renewcommand{\arraystretch}{1.25}
\begin{tabular}{|l|l|l|l|c|}
\hline
\textbf{Topology} & \textbf{Preferred path} & \textbf{Probe orientation} & \textbf{Velocity guidance} & \textbf{Compl.} \\
\hline\hline
Horizontal pipe    & Straight scan, top zenith & Surface-normal (vertical) & $\le$20 mm/s vert.; 10 mm/s horiz. & Low \\
\hline
Vertical pipe      & Lateral plunging path     & Surface-normal, tip to pipe & $\le$30 mm/s, weak $v$-dependence & Med. \\
\hline
Horizontal fitting & Straight pass above nut   & Surface-normal (vertical) & $\le$20 mm/s vert.; 10 mm/s horiz. & Low \\
\hline
Vertical fitting   & Lateral plunging loop     & Surface-normal, tip to nut & $\le$30 mm/s, covers uneven outflow & Med. \\
\hline
\end{tabular}
\end{table*}

\section{Automated Trajectory-Generation Pipeline}
\label{sec_pipeline}
 
The rules of Section~\ref{sec_rules} define what a reliable trajectory must
look like, but applying them by hand to a real piping system is laborious and
error-prone. Because such systems consist largely of repeating features, this
step lends itself to automation. This section presents a proof-of-concept
software pipeline that ingests a digital description of a piping system and
returns a single continuous inspection trajectory suitable for an
augmented-reality assistance system. The goal, in the words of the intended
user, is to provide a piping layout as input and obtain a connected inspection
trajectory as output.
 
\subsection{Architecture and input representation}
\label{sec:pipe:arch}
 
The pipeline mirrors the rule set in software and proceeds in four stages.
First, the components are initialized in a common coordinate frame. Second,
they are ordered into a sensible inspection sequence. Third, a rule-conformant
trajectory is generated for each component. Fourth, the per-component
trajectories are linked into a collision-free global path, which is then
exported as a continuous list of six-DOF inspection poses for an
augmented-reality system.

Rather than parsing native CAD files, the pipeline takes a lightweight
description in which each component is one object with an explicit pose and a
type, plus optional attributes that steer the generation, such as a preferred
generation mode or a probe access direction. This abstracts the assembly into a
machine-readable network of located components while avoiding the burden of
native CAD processing.
 
\subsection{Inspection ordering}
\label{sec:pipe:order}
 
Ordering the components efficiently is posed as a modified Traveling-Salesman
problem in which every component is visited once on the shortest route. Fittings
are compact and represented as single nodes, while elongated pipes are modelled
as nodes with two entry points, so that choosing one pipe end fixes the other as
the start of the next leg. The order is solved either manually or by a greedy
heuristic selecting the nearest open component, into which domain strategies
such as a bottom-up order can be injected.
 
\subsection{Per-component trajectory generation}
\label{sec:pipe:component}
 
For each component the pipeline follows the three classical steps of
computer-aided inspection planning: initialization (gathering type, installed
pose, and any access direction or preference), feature-based segmentation, and
linking. Feature treatment differs by class. Pipes have individual courses, so
their axis course and radius are extracted (the axis course is taken as given
here). Fittings recur in identical form and are handled through a feature
database (Section~\ref{sec:pipe:db}), retrieving sealing-point features by type
rather than re-extracting them per instance.
 
\begin{figure*}[h]
\centering
\includegraphics[width=\linewidth]{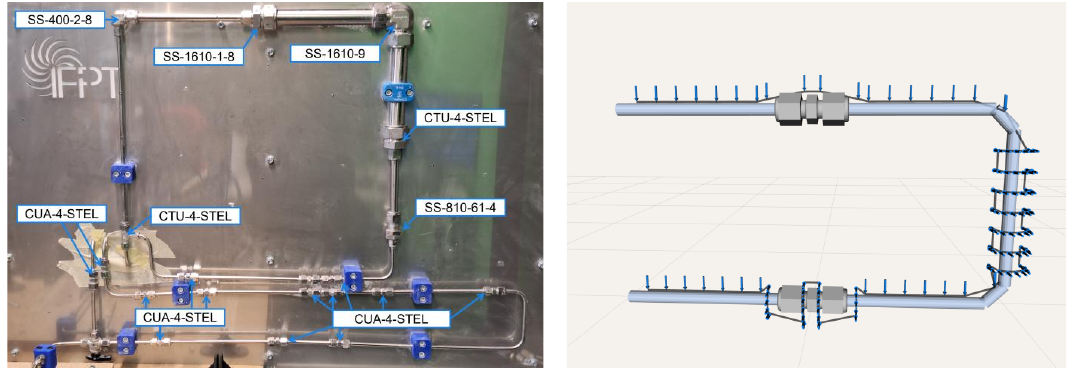}
\caption{Linked segment trajectories on a HyLok CUA-4-STEL fitting. Poses are
placed on each sealing circle, unreachable poses are filtered, and the segments
are ordered into a single component trajectory.}
\label{fig:pipe:fitting_segments}
\end{figure*}

Each feature segment is generated in four steps: poses are distributed along
the surface, unreachable poses are filtered, the rest are ordered for shortest
length, and the segment is assembled. For a fitting, poses are placed on each
sealing circle with orientation set from the access direction
(Figure~\ref{fig:pipe:fitting_segments}); the segments are then linked into a
component trajectory by a small, locally bounded Traveling-Salesman search.
 
Each generated pose carries the spatial coordinates, the surface-normal probe
orientation required by Section~\ref{sec_rules}, and the admissible velocity for
its segment. Once all component trajectories exist, a linking step connects them
into a continuous global trajectory that can be ingested directly by
augmented-reality headsets or collaborative robots to guide the inspection.
 
\subsection{Component feature database}
\label{sec:pipe:db}
 
Because fittings recur in identical or near-identical form, re-analysing each
instance geometrically is inefficient. The pipeline therefore uses a structured
feature database in which each fitting type is described parametrically once.
The data model separates a registry of fitting types from the per-type product
data, so new parts are added without altering the existing schema. Sealing
points are described in a generic, primitive-based form, here a circle defined
by centre, radius, and axis relative to the component origin, kept extensible so
that other primitives can be added later. The requirements that drove this
design are summarized in Table~\ref{tab:pipe:db_req}.
 
\begin{table*}[t]
\centering
\caption{Requirements on the component feature database.}
\label{tab:pipe:db_req}
\setlength{\tabcolsep}{6pt}
\renewcommand{\arraystretch}{1.25}
\begin{tabular}{|l|l|l|}
\hline
\textbf{Category} & \textbf{Requirement} & \textbf{Description} \\
\hline\hline
Functional & Structural flexibility  & Represent the one-to-many relation hierarchically. \\
\hline
Functional & Generic parametrization & Decouple data structure from geometry type. \\
\hline
Functional & Spatial referencing     & Reference all features in one component frame. \\
\hline
Quality    & Scalability             & Add parameters without breaking the structure. \\
\hline
\end{tabular}
\end{table*}
 
\subsection{Validation on virtual piping systems}
\label{sec:pipe:validation}
 
The pipeline was validated on two virtual systems of increasing complexity,
checking classification, per-component planning, transitions, and rule
conformity.
 
In the first case, two pipes joined by a fitting, the pipeline produced a
single continuous trajectory covering each object once: both pipes were
identified as horizontal and assigned a straight line, the CUA-4-STEL features
were retrieved and processed per the specified mode and access direction, and
the output poses were expressed in the local frame with velocity information,
of the kind shown for a single fitting in Figure~\ref{fig:pipe:fitting_segments}.
 
The second case approximates a demonstrator pipe bend with two fittings and
three pipes, one defined in three segments (horizontal, vertical, horizontal)
joined by transitions. All components and segments were classified correctly:
the horizontal segments received straight lines, the vertical segment the
lateral plunging path, the transitions were flagged as neither, and the two
fittings applied their assigned modes (a plunging path and a straight pass).
 
\subsection{Proof-of-concept status and limitations}
\label{sec:pipe:limits}
 
The prototype shows that the rule set can be operationalized reliably, with both
cases producing a complete, rule-conformant trajectory. Two limitations mark the
next steps. First, the pipeline is bounded by the database content and the
quality of the product data, covering only the demonstrator components and
requiring pre-classified components with the pipe-axis course given, so
automated classification of raw geometry is the natural extension. Second, the
validation surfaced two refinements requiring improved pose placement: the
approximated bend transition segments and a residual collision between the
generated poses and the union nuts at one fitting. These points are taken up in
Section~\ref{sec_discussion}.

\section{Discussion}
\label{sec_discussion}
 
The experimental campaign and the pipeline together point to a shift in how
manual hydrogen leak inspection should be specified: intuition alone cannot
reliably detect small leaks under realistic fluid-dynamic conditions, and the
geometry sensitivity reported for large-scale releases~\cite{Kang.2024,
Li.2025} persists in the near field, where it becomes an inspection
requirement rather than a hazard ranking.
 
\subsection{A demonstrated gap in current inspection practice}
\label{sec:disc:practice}
 
The quantified signal degradation exposes a concrete weakness in standard
operating procedures that fix only scalar parameters: the straight scan along
a vertical pipe, executed within every prescribed parameter, detected the
smallest leak in only about 60 percent of passes at the highest tested
velocity (Section~\ref{sec:res:dynamic:vertical}), a false sense of security
rather than a safety verification. The focus on small leaks is deliberate:
large failures produce gas clouds found even with uncoordinated movement,
while in the tens-of-millilitres range, where creeping fatigue failures must
be caught, spatial precision decides. Constraining the path geometry,
enforcing a surface-normal orientation, and coupling the velocity to the
sensor response time restored a high, velocity-tolerant margin, and the
pipeline shows that these constraints can be expressed as machine-readable
trajectories, the prerequisite for deterministic guidance. Complexity,
however, tracks the mounting orientation rather than serving as a general
precaution: where buoyancy concentrates the gas predictably, on horizontal
pipes and fittings, a straight pass with a vertical probe is fast and
adequate, while the plunging path, at roughly fivefold time per sealing
point, is required where the mounting narrows the field or leaves the outflow
direction unpredictable, as on vertical pipes and fittings. Prescribing it
where it adds no detection benefit would not be accepted on the shop floor.
 
\subsection{Designing for an unknown leak}
\label{sec:disc:unknown}
 
A point implicit in the design governs how the rules are applied. At
inspection time the leak has not been found, so its rate, morphology, and
outflow direction are unknown; the only quantity the planner can fix is the
minimum rate that must still be detected, which follows from the system
safety concept~\cite{Masuhr.2024}. A trajectory must therefore be designed
for the least favourable plausible combination of the lowest specified rate
with an unknown geometry and direction. This is why the trajectories were
evaluated at the critical 20 mln/min against the worst observed geometry and
direction per topology: a path that clears the threshold with margin there
also detects larger or more favourably oriented leaks, and it is why a
tangential pass over a fitting is rejected for an encircling one, since only
a path intersecting every radial escape vector guarantees detection when the
direction is unknown.
 
\subsection{What the evidence establishes}
\label{sec:disc:evidence}
 
The added value of this study can be stated in four points. First, a general
safety gap is demonstrated rather than asserted: a scan that satisfies every
scalar requirement of the governing standards still missed the smallest
specified leak in four of ten passes, so the gap lies in the missing spatial
and kinematic specification, not in negligent execution. Second, for the
electrochemical instrument the study yields directly usable trajectories:
Table~\ref{tab:rules:summary} fixes path geometry, probe orientation, and
admissible velocity per topology, validated at the critical leak rate. For the
mass spectrometer, the static maps bound the spatial tolerance a trajectory
must respect and a dynamic check indicates high velocity tolerance, but no
complete rule set is claimed. Third, transfer to further instruments splits into two
parts: velocity limits follow from
re-parameterizing the validated reduction model with a measured step response
(Section~\ref{sec:methods:sensors}), whereas the spatial detectable field must
still be mapped statically, since two aspiration rates do not establish a
scaling law between aspirated flow and field extent; until such data exist,
the two characterized instruments merely bracket the expectation, with
low-aspiration detectors inheriting the more conservative spatial rules.
Fourth, and methodologically most consequential, the dynamic campaign itself
need not be repeated per instrument: the validated first-order model predicts
the dynamic peak from the measured static field and the sensor's $T_{90}$
(Section~\ref{sec:rules:model}), so instrument-specific velocity limits can be
simulated once field and step response are known. Empirical passes remain
necessary only for the position-coupled orientation behaviour and for the
scatter that underlies the margins. The same logic extends across components:
encircling the radial dispersion of a sealing point applies equally to flange
connections and multi-port valves, and since acceptance thresholds are set per
system rather than by harmonized standard~\cite{Masuhr.2024}, the velocity
model provides the system-specific tuning of the inspection speed.
 
\subsection{Deployment as robotic planning or operator assistance}
\label{sec:disc:deployment}
 
The rules and the pipeline support two deployment paths. Executed directly by
a manipulator, the planned velocity and surface-normal orientation are
reproduced exactly, so the velocity factor $F_v$ governs the outcome with
little residual uncertainty; the price is the fixed workspace and the cost of
a robot at every assembly. Visualized through an augmented-reality headset
and executed by a human, the same trajectory scales to arbitrary geometries
without dedicated hardware but adds ergonomics as a planning dimension that
the present complexity rating only approximates. The two are not substitutes:
a skilled operator adapts to the live reading and can compensate residual
registration inaccuracy, which suggests presenting the rule-conformant path
as guidance rather than rigid instruction, so that the algorithm guarantees a
detectable path in outline while human adaptability closes the remaining gap.
 
\subsection{Limitations and future work}
\label{sec:disc:limits}
 
Several limitations bound this study. First, individual static grid points can
misrepresent narrow, directed jets: a small registration offset between probe
tip and leak may miss the jet core, and the electrochemical sensor's
aspiration may partly mask such offsets~\cite{GroeBley.2016}; being artefacts
of single-point sampling, such minima do not carry over to continuous passes,
which accumulate exposure across a spatial neighbourhood. Second, the shielded
bench omits the air currents and obstructing geometry of real installations,
and the support plate below the pipe (Section~\ref{sec:methods:bench}) means
that free downward dispersion was not isolated, so the 180 degree cases
represent installed systems rather than a free-standing pipe. Third, and most
importantly, a robotic manipulator replaced the human operator; the accuracy
of hand-guided execution, and the margins it requires, remain unquantified and
call for a human-subject study. Fourth, the dynamic characterization rests
primarily on the electrochemical instrument (Section~\ref{sec:disc:evidence});
the single-configuration check for the mass spectrometer is not a validation,
and no static campaign exists yet for a third instrument. Fifth, the pipeline
requires pre-classified components and covers only the demonstrator database.
Finally, five repetitions per condition reveal the strong effects reported but
preclude formal confidence intervals.
 
\section{Conclusions}
\label{sec_conclusion}
 
Ensuring the hermetic integrity of hydrogen infrastructure is, in electrolyzer
manufacturing, verified almost entirely by manual tracer-gas sniffing. This
study quantified the process-induced uncertainty of that procedure and exposed
a gap in the governing standards, which mandate leak testing but specify no
spatial routing of the probe. Using a robotically guided test bench, it showed
that scanning velocity and position-coupled probe orientation govern
detectability: a straight scan along a vertical pipe missed the smallest leak
in four of ten passes at the highest tested velocity, whereas encircling paths
maintained a high margin throughout. From these observations a physics-based
rule set was derived, comprising a velocity-dependent reduction model tying
scanning speed to the sensor response time, a position-coupled orientation
rule, and topology-specific routing rules, and operationalized in a
proof-of-concept pipeline that generates rule-conformant trajectories from
structured 3D data for assistance systems. Hydrogen safety guidelines should
evolve to include explicit spatial and kinematic routing requirements, not
only scalar parameters. Validating the trajectories with human operators and
extending the planning to unlabelled geometry are the next steps toward
deterministic, assistance-guided inspection.



\bibliographystyle{unsrturl}
\bibliography{bibliography}

\end{document}